\documentclass[10pt, a4paper]{article}
\usepackage{lrec2016}
\usepackage[utf8]{inputenc}
\usepackage[T2A,T1]{fontenc}
\usepackage[russian,english]{babel}
\usepackage{multibib}
\newcites{languageresource}{Language Resources}
\usepackage{graphicx}
\usepackage{epstopdf}
\usepackage{amssymb}
\usepackage{amsmath}
\usepackage{amsfonts}
\usepackage{hyperref}
\usepackage{array,tabularx,tabulary,booktabs}
\usepackage{multirow}
\usepackage{blindtext}
\usepackage{pbox}

\title{Clustering Comparable Corpora of Russian and Ukrainian Academic Texts: Word Embeddings and Semantic Fingerprints} 

\name{Andrey Kutuzov\textsuperscript{1}, Mikhail Kopotev\textsuperscript{2}, Tatyana Sviridenko\textsuperscript{3}, Lyubov Ivanova\textsuperscript{3}}

\address{University of Oslo\textsuperscript{1}, University of Helsinki\textsuperscript{2}, National Research University Higher School of Economics\textsuperscript{3}\\
         Oslo\textsuperscript{1}, Helsinki\textsuperscript{2}, Moscow\textsuperscript{3} \\
         andreku@ifi.uio.no, mihail.kopotev@helsinki.fi, sviridenkot@gmail.com, luben92@gmail.com\\}

\abstract{We present our experience in applying distributional semantics (neural word embeddings) to the problem of representing and clustering documents in a bilingual comparable corpus. Our data is a collection of Russian and Ukrainian academic texts, for which topics are their academic fields. In order to build language-independent semantic representations of these documents, we train neural distributional models on monolingual corpora and learn the optimal linear transformation of vectors from one language to another. The resulting vectors are then used to produce `semantic fingerprints' of documents, serving as input to a clustering algorithm. 
The presented method is compared to several baselines including `orthographic translation' with Levenshtein edit distance and outperforms them by a large margin. We also show that language-independent `semantic fingerprints' are superior to multi-lingual clustering algorithms proposed in the previous work, at the same time requiring less linguistic resources.\\
\newline 
 }

\begin{document}

\maketitleabstract

\section{Introduction}\label{sec:intro}
This research addresses the problem of representing the semantics of text documents in multi-lingual comparable corpora. We present a new approach to this problem, based on neural embeddings, and test it on the task of clustering texts into meaningful classes depending on their topics. The setting is unsupervised, meaning that one either does not have enough annotated data to train a supervised classifier or does not want to be limited with a pre-defined set of classes. There is a lot of sufficiently good approaches to this problem in the case of mono-lingual text collections, but the presence of multiple languages introduces complications. 

When a text collection contains documents in several languages, it becomes impractical to simply represent the documents as vectors of words occurring in them ("bag-of-words"), as the words surface forms are different, even in closely-related languages. Thus, one has to invent means to cross the inter-lingual gap and bring all documents to some sort of shared representation, without losing information about their topics or categories.

Of course, one obvious way to solve this problem is to translate all documents into one language, and then apply any clustering algorithm. However, this requires either buying human/machine translation services (which can be expensive if you deal with large text collection) or training own statistical machine translation model (which as a rule requires big parallel corpus). This is the reason to search for other solutions.

In this paper, a novel way of reducing the problem of cross-lingual document representation to a monolingual setting is proposed. Essentially, we train Continuous Bag-of-Words models \cite{Mikolov_representation:2013} on large comparable monolingual corpora for two languages our dataset consists of. This provides us with vector representations of words, allowing to measure their semantic similarity. Then, a linear transformation matrix from vectors of language \textit{A} to vectors of language \textit{B} is learned, using a small bilingual dictionary as training data. This matrix is then employed to `project' word and document representations from semantic space of language \textit{A} to semantic space of language \textit{B}. It allows not only quite accurate `translation' of words, but also of document `\textit{semantic fingerprints}' (dense representations of document semantics, calculated as an average of the trained distributional vectors for all the words in document).

This approach is evaluated in a setting, where the input is a collection of documents in several languages and some number of topics to which these documents belong (we also have large monolingual corpora to train distributional models on). For each document, we are given its language, but not its topic. The task is to cluster this collection so that documents belonging to one topic were clustered together, independent of their language. Note that we are interested in clustering the collection as a whole, not each language separately (which is trivial).

Our evaluation data consists of comparable corpora of Russian and Ukrainian academic texts. On this material, we show that the `\textit{translated semantic fingerprints}' method represents documents in different languages precisely enough to allow almost exact clustering according to document topics, with only 5\% of incorrect assignments. It significantly outperforms both naive bag-of-words baseline and the not-so-naive method of `orthographic translation' based on Damerau-Levenshtein distance, even enriched with dictionary mappings. At the same time, it does not require large parallel corpora or a ready-made statistical machine translation model.

The rest of the paper is structured as follows. In Section \ref{sec:related} we describe the foundations of our approach and the related work. Section \ref{sec:data} introduces the employed corpora and the story behind them. Section \ref{sec:matrix} is dedicated to learning the transformation matrix, and Section \ref{sec:experiment} describes our experimental setting and evaluation results. We discuss the findings in Section \ref{sec:disc} and conclude in Section \ref{sec:conclusion}, also suggesting directions for future work.

\section{Related Work}\label{sec:related}
Clustering multi-lingual documents has received much attention in natural language processing. Among approaches not using some form of machine translation, one can mention \cite{mathieu2004multilingual}, who essentially employ a bilingual dictionary to bring some words in the documents to a language-independent form and then to perform clustering. In the section \ref{sec:experiment} we show that our approach based on neural embeddings significantly outperforms their reported results.

\cite{wolf2014joint} proposed training joint multi-lingual neural embedding models. Theoretically, this can be used to achieve our aim of language-independent semantic representations for documents. Unfortunately, it demands a large word-aligned parallel corpus. This is not the case with the more recent \textit{Trans-gram} approach introduced in \cite{coulmance2016trans}, also able to learn multi-lingual models. However, it still needs sentence-aligned corpora to train on (in the size of millions of paired sentences).  Large parallel corpora (whether word- or sentence-aligned) are often a scarce resource, especially in the case of under-represented languages. 

The approach described in this paper takes as an input only comparable monolingual corpora and bilingual dictionaries in the size of several thousand word pairs. Such resources are much easier to find and evaluate. We employ the idea of learning a linear transformation matrix to map or project word embeddings from the semantic space of one language to that of another. This idea was first proposed in \cite{mikolov2013translation}, who applied it to lexical translation between English, Spanish, Czech and Vietnamese. We extend it from continuous representations of single words or collocations to `\textit{semantic fingerprints}' of documents as a whole.

\section{Academic texts as Comparable Corpora}\label{sec:data}
The Russian and Ukrainian languages are mainly spoken in Russian Federation and the Ukraine and belong to the East-Slavic group of the Indo-European language family. They share many common morphosyntactic features: both are SVO languages with free word order and rich morphology, both use the Cyrillic alphabet and share many common cognates. 

Both Russia and the Ukraine have common academic tradition that makes it easier to collect corpora, which are comparable in terms of both genre and strictly defined academic fields. We work with such a corpus of Russian and Ukrainian academic texts, initially collected for the purposes of cross-lingual plagiarism detection. This data is available online through a number of library services, but unfortunately cannot be republished due to copyright limitations.

The Ukrainian subcorpus contains about 60 thousand extended summaries (Russian and Ukrainian \foreignlanguage{russian}{`автореферат'}, `\textit{avtoreferat}') of theses submitted between 1998 and 2011. The Russian subcorpus is smaller in the number of documents (about 16 thousand, approximately the same time period), but the documents are full texts of theses, thus the total volume of the Russian subcorpus is notably larger: 830 million tokens versus 250 million tokens in the Ukrainian one. Generally, the texts belong to one genre that can be defined as post-Soviet expository academic prose, submitted for academic degree award process.

The documents were converted to plain text files from MS Word format in the case of the Ukrainian subcorpus and mainly from OCRed PDF files in the case of the Russian subcorpus. Because of this, the Russian documents often suffer from OCR artifacts, such as words split with line breaks, incorrectly recognized characters and so on. However, it does not influence the resulting model much, as we show below.

Both Ukrainian and Russian documents come with meta data allowing to separate them into academic fields, with economics, medicine and law being most frequent topics for the Ukrainian data and economics, history and pedagogy dominating the Russian data.

For evaluation, 3 topics were used, distant enough from each other and abundantly presented in both subcorpora: economics, law and history. We randomly selected 100 texts in each language for each topic. As an average length of Russian texts is significantly higher (them being full theses), we cropped them, leaving only the first 5 thousand words, to mimic the size of the Ukrainian summaries. These 600 documents in 3 classes are used as a test set (see Section \ref{sec:experiment} for the description of the conducted experiments).

The corpora (including test set) were PoS-tagged\footnote{We used \textit{Mystem} \cite{Segalovich:2003} for Russian and \textit{Ugtag} \cite{kotsyba2009ugtag} for Ukrainian.}. Each word was replaced with its lemma followed by a PoS-tag (`\foreignlanguage{russian}{диссертация\_S}', `\foreignlanguage{russian}{диссертацiя\_N}'). Functional parts of speech (conjunctions, pronouns, prepositions, etc.) and numerals were removed from the texts.

\section{Learning to Translate: Ukrainian-to-Russian transformations}\label{sec:matrix}
As already stated, our main proposal is using neural embedding models to `project' documents in one language into the semantic space of another language. For this, we first trained a Continuous Bag-of-Words (CBOW) and a Continuous SkipGram model \cite{Mikolov_representation:2013} for each of our monolingual subcorpora. The models were trained with identical hyperparameters: vector size of 300 components\footnote{\cite{mikolov2013translation} suggest to use larger vector size for the source language model; however, we leave it for the future work.}, symmetric window of 2 words, negative sampling with 10 samples, 5 iterations over the corpus, no down-sampling. The only language-dependent difference was that for the Ukrainian model we ignored words with the corpus frequency less than 10 and for the Russian model this threshold was set to 15 (as the Russian corpus is 3 times larger). All in all, the final Ukrainian model recognizes 429 215 words and the Russian one 271 720 words. Training was performed using CBOW and SkipGram implementation in \textit{Gensim} library \cite{Rehurek2010gensim}.

After the models were trained, we followed the path outlined in \cite{mikolov2013translation} to learn a linear transformation matrix from Ukrainian to Russian. First, we extracted all noun pairs from Russian-Ukrainian bilingual dictionary \cite{ganich1990}, with the constraint that their frequency in our corpora was above the already mentioned thresholds 15 and 10 for Russian and Ukrainian words correspondingly. That made it a list of about 5 thousand pairs of nouns being translations of each other. 

For all these words, their vectors were found in the models corresponding to the words' languages. It provided us with a matrix of 5 thousand of 300-dimensional Ukrainian vectors and the matrix of corresponding 5 thousand of 300-dimensional Russian vectors. This data served as a training set to learn an optimal  transformation matrix. The latter is actually a 300x301 matrix of coefficients, such that when the initial Ukrainian matrix is multiplied by this transformation matrix, the result is maximally close to the corresponding Russian matrix. This transformation matrix has 301 (not 300) columns, because we add one component equal to 1 to each vector, as a bias term.  

Producing the transformation matrix is a linear regression problem: the input is 301 components of Ukrainian vectors (including the bias term) and the output is 300 components of Russian vectors. As we need 300 values as an output, there are actually 300 linear regression problems and that's why the resulting matrix size is 300x301 (301 weights for each of 300 components).

There are two main ways to solve a linear regression problem: one can either learn the optimal weights in an iterative way using some variant of gradient descent, or one can solve it numerically without iteration, using normal equation. For English and Spanish, \cite{mikolov2013translation} used stochastic gradient descent. However, normal equation is actually less error-prone and is guaranteed to find the global optimum. Its only disadvantage is that it becomes very computationally expensive when the number of features is large (thousands and more). However, in our case the number of features is only 301, so computational complexity is not an issue.

Thus, we use normal equation to find the optimal transformation matrix. The algebraic solution to each of 300 normal equations (one for each vector component $i$) is shown in the Equation \ref{eq:normal}:
\begin{equation}\label{eq:normal}
\boldsymbol\beta_i = (\textbf{X}^\intercal* \textbf{X})^{-1} * \textbf{X}^\intercal * y_i
\end{equation}
where $\textbf{X}$ is the matrix of 5 thousand Ukrainian word vectors (input), $y_i$ is the vector of the $i$th components of 5 thousand corresponding Russian words (correct predictions), and $\boldsymbol\beta_i$ is our aim: the vector of 301 optimal coefficients which transform the Ukrainian vectors into the $i$th component of the Russian vectors. 

After solving such normal equations for all the 300 components $i$, we have the 300x301 linear transformation matrix which fits the data best.

This matrix basically maps the Ukrainian vectors into the Russian ones. It is based on the assumption that the relations between semantic concepts in different languages are in fact very similar (\textit{students} are close to \textit{teachers}, while \textit{pirates} are close to \textit{corsairs}, and so on). In continuous distributional models which strive to represent these semantic spaces, mutual `geometrical' relations between vectors representing particular words are also similar across models (if they are trained on comparable corpora), but the exact vectors for words denoting one and the same notion are different. This is because the models themselves are stochastic and the particular values of vectors (unlike their positions in relation to each other) depend a lot on technical factors, including the random seed used to initialize vectors prior to training. In order to migrate from a model \textbf{A} to another model \textbf{B}, one has to `rotate and scale' \textbf{A} vectors in a uniform linear way. To learn the optimal transformation matrix means to find out the exact directions of rotating and scaling, which minimize prediction errors.

Linguistically speaking, once we learned the transformation matrix, we can predict what a Russian vector would most probably be, given a Ukrainian one. This essentially means we are able to `translate' Ukrainian words into Russian, by calculating the word in the Russian model with the vector closest to the predicted one.

We had to choose between CBOW or Continuous SkipGram models to use when learning the transformation matrix. Also, there was a question of whether to employ regularized or standard normal equations. Regularization is an attempt to avoid over-fitting by trying to somehow decrease the values of learned weights. The regularized normal equation is shown in \ref{eq:normalreg}:

\begin{equation}\label{eq:normalreg}
\boldsymbol\beta_i = (\textbf{X}^\intercal* \textbf{X} + \lambda * L)^{-1} * \textbf{X}^\intercal * y_i
\end{equation}

Comparing to \ref{eq:normal}, it adds the term $\lambda * L$, where $L$ is the identity matrix of the size equal to the number of features, with 0 at the top left cell, and $\lambda$ is a real number used to tune the influence of regularization term (if $\lambda = 0$, there is no regularization).

To test all the possible combinations of parameters, we divided the bilingual dictionary into 4500 noun pairs used as a training set and 500 noun pairs used as a test set. We then learned transformation matrices on the training set using both training algorithms (CBOW and SkipGram) and several values of regularization $\lambda$ from 0 to 5, with a step of 0.5. The resulting matrices were applied to the Ukrainian vectors from the test set and the corresponding Russian `translations' were calculated. The ratio of correct `translations' (matches) was used as an evaluation measure. It came out that regularization only worsened the results for both algorithms, so in the Table \ref{tab:matrix} we report the results without regularization. 

For reference, we also report the accuracy of `quazi-translation' via Damerau-Levenshtein edit distance \cite{Damerau:1964}, as a sort of a baseline. As already stated, the two languages share many cognates, and a lot of Ukrainian words can be orthographically transformed into their Russian  translations (and vice versa) by one or two character replacements. Thus, we extracted 50,000 most frequent nouns from our Russian corpora; then for each Ukrainian noun in the bilingual dictionary we found the closest Russian noun (or 5 closest nouns for @5 metric) by edit distance and calculated how often it turned out to be the correct translation. As the Table \ref{tab:matrix} shows, notwithstanding the orthographic similarity of the two languages, CBOW consistently outperforms this approach even on the test set. On the training set, its superiority is even more obvious.

\begin{table}
 \caption{Translation accuracy}\label{tab:matrix}
\begin{tabular}{lcc|cc|l}
\toprule
& \multicolumn{2}{c|}{\textbf{CBOW}}  & \multicolumn{2}{c|}{\textbf{SkipGram}} & \parbox[c]{2px}{\textbf{Edit} \\\textbf{distance}}\\ 
& Training&Test&Training&Test&\\ 
\midrule
\textbf{@1} & 0.648 & \textbf{0.57}& 0.545&0.374&0.549\\   
\textbf{@5} & 0.764 & \textbf{0.658} & 0.644 &0.486&0.619 \\
\midrule
\end{tabular}
\end{table}

As for comparison between learning algorithms for matrix translation, CBOW-based transformation matrix is again the winner, with 57\% matches on the test set and 65\% matches on the training set, beating SkipGram in both. Note that in the context of this task, SkipGram models seem to have problems with actually learning the optimal transformation matrix for unseen data: on the test set they perform even worse than the edit distance approach.

CBOW is also consistently better if we consider cases when the correct word is among 5 nearest neighbors of the predicted vector to be matches as well (accuracy @5). This is an important metrics, because quite often the `translation' is not exactly the corresponding word from the dictionary, but still a very semantically similar one, while the dictionary translation is the second or the third by its cosine similarity to the predicted vector. It means that in fact the `semantic translation' is successful, as the concept is correct. For example, our algorithm translates the Ukrainian noun `\foreignlanguage{russian}{гетьман}' \textit{hetman} into Russian `\foreignlanguage{russian}{царь}' \textit{tzar}, while the correct translation `\foreignlanguage{russian}{гетман}' is the second nearest neighbor. 

Notwithstanding the fact that the transformation matrix was trained exclusively on nouns, it correctly `translates' adjectives and verbs as well (we did not experiment with other parts of speech though). However, it tends to `substantivize' them: for example, the Ukrainian verb `\foreignlanguage{russian}{розробити}' \textit{to develop} is transformed into a Russian vector, which is closer to the noun `\foreignlanguage{russian}{разработка}' \textit{development} than to the corresponding verb. 

Thus, at least main parts of speech seem to share a common Ukrainian-to-Russian projection matrix, supporting the view that semantic spaces for different languages are in comparatively simple linear relations to each other. In the following clustering experiments we employed CBOW-based transformation matrix and consequently CBOW models for Russian and Ukrainian.

We also applied the same transformation matrix to the document-level `semantic fingerprints'. These fingerprints are simple average vectors of all words that the document contains. Thus, if our models have vector size 300, the resulting fingerprints are 300-dimensional vectors as well. These vectors can be transformed with the same matrix. As we show in the Section \ref{sec:experiment}, the cross-lingual linear relations hold not only for words, but for these semantic fingerprints as well. 

\section{Experiment Design and Evaluation}\label{sec:experiment}
We evaluate the cross-lingual representations described above on the task of clustering a set of documents. Recall that our test set consists of randomly selected 600 documents, equally divided between the topics of economics, law and history, and Russian and Ukrainian languages. Thus, we have 100 Ukrainian law texts, 100 Russian law texts, etc. The average length of the texts is 4000 word tokens.

We aim to find such a representation for documents which would reveal their topical structure independent of the language. It can be tested by clustering the whole collection in an unsupervised way into 3 clusters (in our setting, the number of topics is a given parameter), and finding out to what extent these clusters correspond to the topical classes: law, economics and history. This correspondence can be calculated by mapping the resulting clusters into topics judging by where the majority of documents belonging to this or that topic were assigned. For example, if more than 100 history documents were assigned to the cluster 0, we map this cluster to the history topic, etc. Then, the ratio of incorrect assignments is calculated, as percentage from the total number of documents. This is our primary evaluation measure. All the clustering experiments below are performed using a well-established \textit{K-means} algorithm \cite{hartigan1975clustering} with Euclidean distances. We intentionally employ the most basic clustering algorithm to make the difference of the underlying representations more visible.

The lemmatized documents were represented as bags-of-words. To reduce the dimensionality of such representations and to filter out unimportant noise words, some sort of feature selection is often used. We employed the most basic variant of it: frequency threshold, where the words are ranked by their frequencies in the whole document collection, and only top $x$ are then used in constructing vector representations. We empirically chose $x=500$, as several values from 100 to 1000 which we tried (with the step of 100) resulted in worse performance, independent of the approaches tested. Note that the sets of 500 most frequent words were selected for each topic separately, to avoid the situation when some topics are under-represented, because words related to them are not frequent. Then the union of these sets was used as the final vocabulary (resulting in vectors of about 800...900 dimensions, depending on the particular method used). Initially, binary vectors were constructed (a word is either present in the document or not), but we also tested count vectors, which store words' per-document frequencies; see below.

In order to make sure that the topical division is indeed manifested in the documents, we first clustered Ukrainian and Russian corpora separately, using the binary bag-of-words representations described above. This gave only 4.7\% incorrect assignments for the Ukrainian texts and 34.7\% incorrect assignments for the Russian part of the test set. Thus, for Ukrainian the division is almost perfect, while for Russian it is manifested less clearly (it seems that economics and law are consistently mixed up), but still the overwhelming majority of documents is clustered according to the topics. It means that the test set does contain information to correctly cluster the documents on a mono-lingual level, and it makes sense to try to achieve comparable (or at least not much worse) results in the cross-lingual experiments. Note that one can't simply cluster the documents in Russian and in Ukrainian separately to achieve our aim: even if the clusterings are ideal, there will be no way to map the Russian clusters to the Ukrainian ones, or vice versa.

So, the next step was to cluster all documents together, independent of their languages, using the techniques described in the Section \ref{sec:matrix}. The results are shown in the Table \ref{tab:evaluation}. 

We used two simple baseline approaches. The first one is dubbed `\textbf{naive}': we cluster all the texts' bag-of-words representations as is, with no special preprocessing (only the PoS tags are unified across languages). Transformation from texts to bags-of-words resulted in 885-dimensional document vectors. This baseline approach exploits the intuition that in closely-related languages such as Russian and the Ukrainian there are many words which share both spelling and meaning. This is true, but this fact does not help \textit{K-means} to correctly cluster the collection into topical classes: 50.17\% of the documents are assigned an incorrect cluster, much more than in any of our mono-lingual experiments. Employing count vectors instead of binary ones lowers error rate only down to 50\%. Using \textit{tf-idf} weighting \cite{sparck1972} did not significantly change the results neither for this nor for other baselines.

Looking into particular cluster assignments reveals that \textit{K-means} clusters all the Ukrainian documents into one group, and then partitions Russian texts into two clusters roughly corresponding to history and everything else. This is quite expected: the Ukrainian alphabet contains several frequent characters missing in Russian (\foreignlanguage{russian}{`ґ, є, і, ї'}), while the Russian-specific characters (\foreignlanguage{russian}{`ё, ъ, ы, э'}) are much rarer. Consequently, the Ukrainian documents contain a lot of Ukrainian words specific only to them, while Russian words (or their identically spelled Ukrainian counterparts) are used throughout the whole collection. Anyhow, `\textbf{naive}' approach can't adequately represent the topical structure of the test set.

\begin{table}
 \caption{Clustering correspondence to document topics}\label{tab:evaluation}
\begin{tabular}{lc}
\toprule
\textbf{Method}&\parbox[c]{80px}{\textbf{Incorrect assignments, \%}} \\  \midrule
\multicolumn{2}{c}{\textbf{Mono-lingual}}   \\
Ukrainian & 4.7 \\
Russian & 34.7 \\
\midrule
\multicolumn{2}{c}{\textbf{Cross-lingual}} \\
Naive Binary &  50.17\\   
Naive Count & 50.00  \\
\midrule
Edit distance translation Binary & 50.50 \\   
Edit distance translation Count & 50.50  \\
\midrule
Dictionary/Edit distance Binary & 50.33 \\   
Dictionary/Edit distance Count & 49.83  \\
\midrule
Matrix translation Binary & 36.33 \\   
Matrix translation Count & 36.17  \\
\midrule
Semantic fingerprints on word types & 35.33  \\
Semantic fingerprints on word tokens & \textbf{5.50}  \\
\midrule
\end{tabular}
\end{table}

The second baseline employs quazi-translation of Ukrainian words into Russian using the already described approach with Damerau-Levenshtein edit distance \cite{Damerau:1964}. We replaced all words in the Ukrainian texts with the Russian words closest to them by edit distance. Only nouns, adjectives, verbs and abbreviations were replaced; replacements were selected only among the same part of speech as the original word, and in case of ties, target word with the highest frequency in Russian corpus was selected. Then the same bag-of-words representation (now 834-dimensional) was fed to the clustering algorithm. Though for many words `\textbf{edit distance translation}' works quite well, it did not help in clustering multilingual test set.  Whether with binary or count vectors, \textit{K-means} still grouped all the Ukrainian documents into one cluster, resulting in 50.5\% of incorrect assignments. The possible reason is that there are still many incorrect `Levenshtein translations' resulting in target entities which are correct Russian words, but never appear in Russian documents from our test set. This gives \textit{K-means} the ground to separate the Ukrainian texts from all the other documents.

Then we experimented with translating Ukrainian words into Russian using the learned transformation matrix (\textbf{matrix translation}). For each Ukrainian word, we multiplied its vector in the Ukrainian model by the matrix and found the Russian word nearest to the resulting vector. Then the Ukrainian words were replaced with these `translations' and the same bag-of-words document representations were constructed (resulting in 845-dimensional vectors). As a result, the \textit{K-means} clustering moved substantially towards the intended topical grouping: only 36.33\% of the documents were assigned incorrect clusters, and using count vectors made it 36.17\%. In fact, the clustering algorithm correctly separated all history documents into one cluster independent of the language, while still mixing things up with law and economics (as we know, they are a bit more difficult to separate even in a monolingual setting). Thus, this document representation seems to be clearly superior to the baseline \textbf{naive} or \textbf{edit distance} approaches. It results in the documents grouping which is almost as efficient as mono-lingual clustering of Russian texts, but is still not on a par with Ukrainian mono-lingual clustering.

Note that these improvements cannot be explained by the sheer fact of employing a bilingual dictionary. We tried to use the same dictionary directly: that is, for the Ukrainian texts in the test set, replace all the words with their dictionary Russian translation. The remaining out-of-vocabulary words were `translated' with the Dameral-Levenshtein distance approach. The results are reported in the Table \ref{tab:evaluation} as \textbf{dictionary/edit distance} method. They are a bit better than the ones of the raw edit distance, but still far from the performance of the \textbf{matrix translation} method. It means that the algorithm itself is the cause of improvements.

Finally, the best results were received by employing the `\textbf{semantic fingerprint}' approach. Recall that this fingerprint is an average vector of all words in the document. Consequently, each document is represented with a 300-dimensional vector, supposedly reflecting its `meaning'. 

The average vector can be calculated either on vectors of \textbf{word types} or of \textbf{word tokens} (thus taking into account individual frequencies of words in the document). These two variants roughly correspond to \textbf{binary} and \textbf{count} variants of the previous methods, but we intentionally dub them in another way to emphasize that these representations are radically different from bags-of-words. In this case we abstract away from particular words, and instead use some generalized `semantic components', hopefully similar across languages.

We computed such fingerprints for all the documents in the test set, and for the Ukrainian documents we then multiplied the fingerprints by the transformation matrix, thus `projecting' them into Russian semantic space. The resulting 300-dimensional representations are already numerical and can be directly fed into a clustering algorithm, without any bag-of-words preprocessing.

As a result, even rough semantic fingerprints calculated on word types (on \textbf{sets} of words in the documents) show clustering accuracy 1\% better than the \textbf{matrix translation} approach. But as soon as semantic fingerprints are computed using word tokens (\textbf{lists} of words), the ratio of incorrect assignments drops drastically down to 5.5\%. This result is very close to the quality of the mono-lingual Ukrainian clustering and is much better than that of the mono-lingual Russian clustering. It means that semantic fingerprints approach performs almost as good in the cross-lingual setting as traditional approaches in the mono-lingual one. Additionally, the fact that it outperformed the Russian mono-lingual clustering might mean that using dense vector representations for documents allowed to overcome the problems with separating economics and law texts in Russian, which seemed intractable for the bag-of-words approach.

To be more precise, into their respective clusters were grouped 196 of 200 economics documents (this corresponds to approximately 0.95 precision and 0.98 recall), 195 of 200 history documents (0.92 precision, 0.98 recall) and 176 of 200 law documents (0.97 precision, 0.88 recall), all independent of their languages. Total average F1 measure is about 0.95, which significantly outperforms the multilingual clustering performance reported in \cite{mathieu2004multilingual}.

\begin{figure}
  \includegraphics[width=\linewidth]{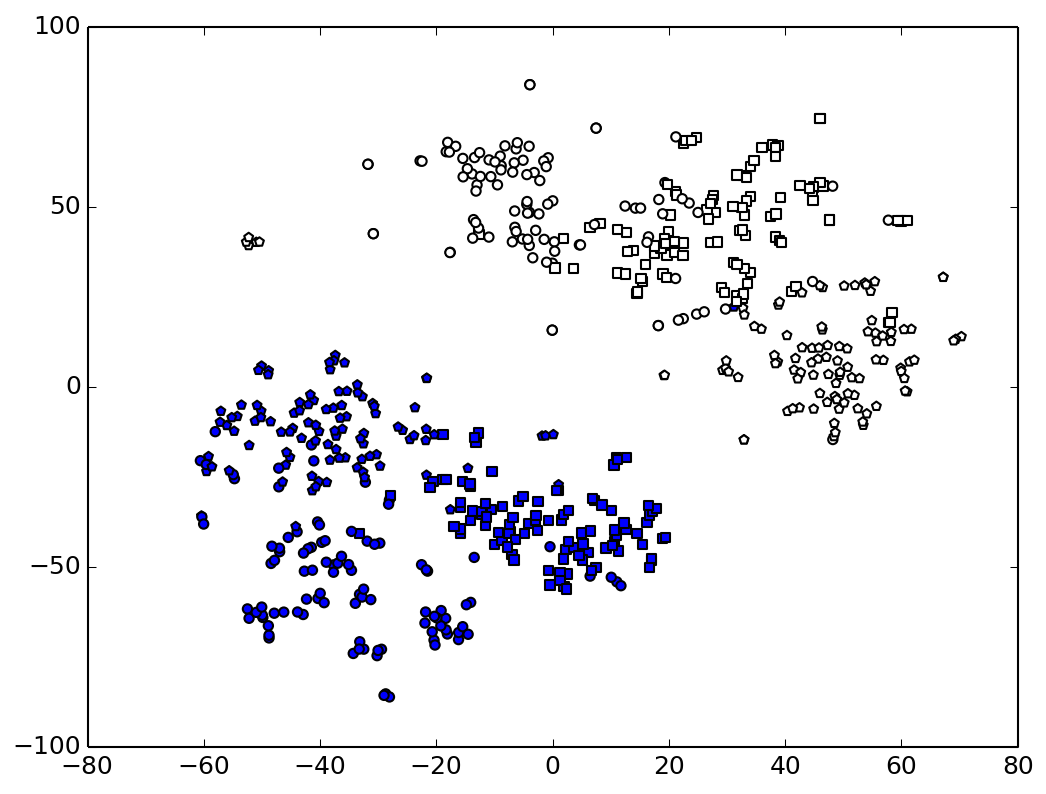}
  \caption{Naive baseline clustering}\label{fig:baseline}
\end{figure}

\begin{figure}
  \includegraphics[width=\linewidth]{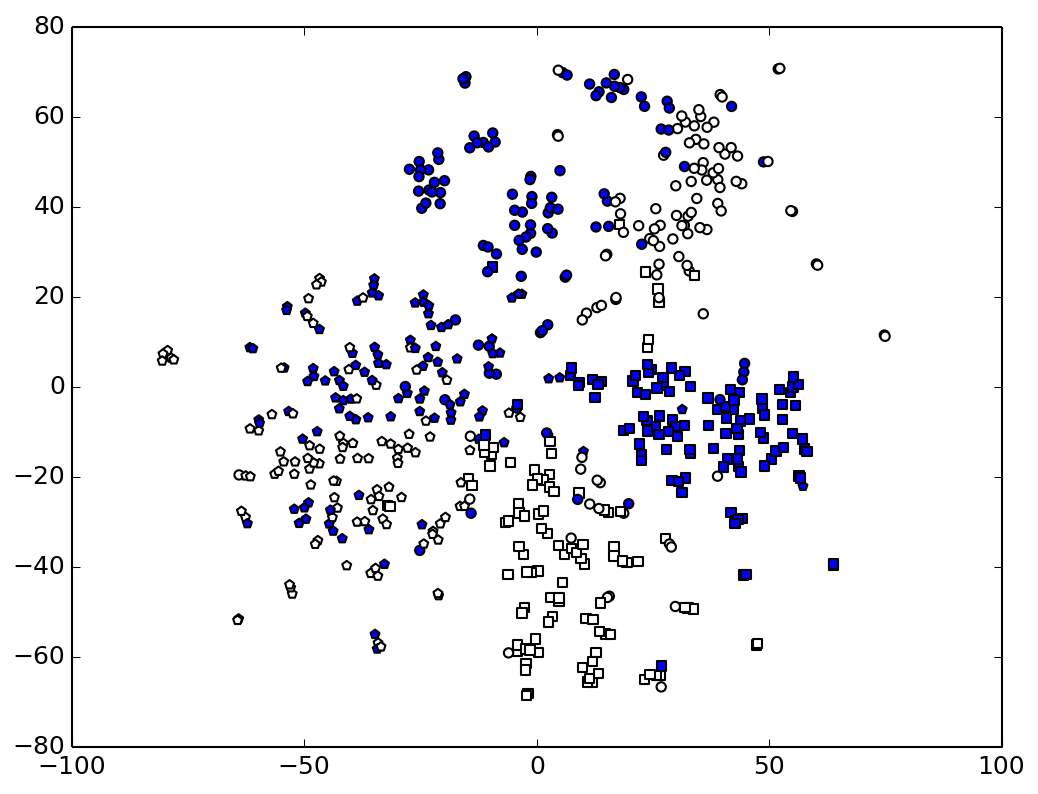}
  \caption{Matrix translation clustering}\label{fig:matrix}
\end{figure}

\begin{figure}
  \includegraphics[width=\linewidth]{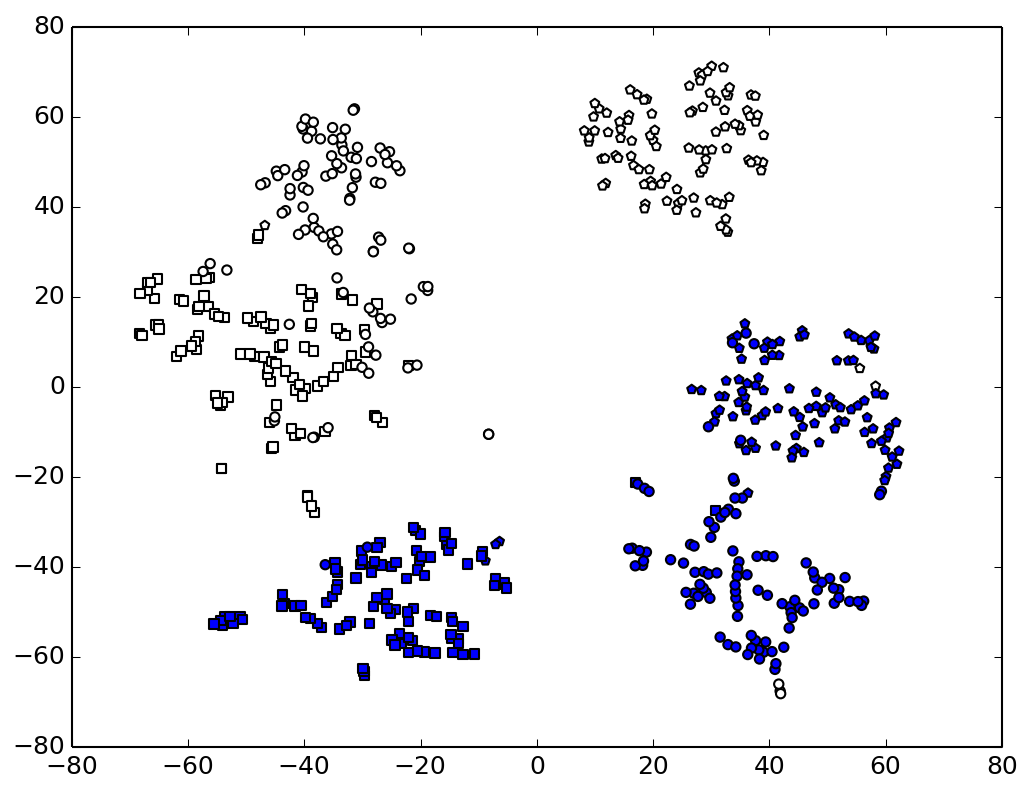}
  \caption{Semantic fingerprints clustering}\label{fig:fingerprint}
\end{figure}

Figures \ref{fig:baseline}, \ref{fig:matrix} and \ref{fig:fingerprint} illustrate clustering mechanics for the methods described above. We employed \textit{t-SNE} dimensionality reduction technique \cite{Maaten:2008} to project high-dimensional representations\footnote{300 dimensions for semantic fingerprints, 885 and 845 for naive baseline and matrix translation correspondingly.} of the test set documents into 2-dimensional plots. Colors reflect document language (blue for Ukrainian and white for Russian), while marker types stand for document topic (circles for law, squares for history and pentagons for economics). Note that these projections inevitably lose a lot of information as compared to initial high-dimensional data, and should be considered as only approximate visualizations.

It is clearly visible that with \textbf{naive baseline} representations in the Figure \ref{fig:baseline} there are almost no links between different-language documents belonging to one topic. The dataset is clearly separated into Russian and Ukrainian clusters, and topics can be seen inside languages, but there is hardly a way to group documents into language-independent topical clusters. This is the reason for \textit{K-means} failing to achieve our aim with the baseline approach.  On the other hand, with \textbf{matrix translation} representations (Figure \ref{fig:matrix}), language-independent topics already emerge, but still with much noise. Language boundaries are eroded, especially with economics documents.

Finally, with \textbf{semantic fingerprints} representations in the Figure \ref{fig:fingerprint} the structure of the test set is manifested in full. There are six well-defined clusters corresponding to topics and languages and a clear spatial structure, which allows \textit{K-means} to easily group documents into 3 larger topical clusters without losing the ability to tell a Russian document from a Ukrainian one. Note how the Ukrainian topical clusters seem to share a common linear relation to the Russian ones, reminding about linear relations between different languages' vector spaces.

Thus, we were able to correctly cluster multilingual documents according to their topics without any proper `translation' and without even considering word spelling. This means that, first, semantic fingerprints are precise enough to reveal topical differences between documents, and second, that this holds even after linear transformation of such fingerprints into another language semantic space.

\section{Discussion}\label{sec:disc}
We tested `transformed' semantic representations of the documents on the clustering task, but theoretically they can be used for any problems which demand semantic-aware cross-lingual representations, including classification and visualization. Also, the number of involved languages is not limited in any way. The proposed method is relatively simple and straightforward to implement: one needs only comparable monolingual corpora to train CBOW models on them (using any of the available off-the-shelf toolkits) and a small bilingual dictionary for each language pair to train linear transformation matrices. After that, all the words and documents in the corpora can be transformed into a unified language-independent semantic representation.

It is interesting that in our experiments semantic fingerprints' performance was better than direct `translation' of words using the same transformation matrix. It reveals an important advantage of such generalized representations: they do not depend on particular words. In the case of the bag-of-words approach, a small mistake in matrix translation can lead to replacement of a Ukrainian word with a Russian counterpart, which is semantically similar, but not exactly the one used to denote this concept in the Russian part of this text collection. As a result, this word becomes useless in representing documents cross-linguistically. On the other hand, with the semantic fingerprints approach, an `approximate transformation' is enough, as it will still be close to the corresponding words from Russian texts in the vector space. 

This also explains the accuracy boost this approach gets from considering word tokens instead of word types. Of course, one reason is that vectors for frequent (arguably more topical) words become more important in determining the final average value, but this is also the case for bag-of-words approaches. The difference is that with the latter (including `matrix translation' method), frequencies of words from the same semantic field are interpreted as independent features. As a result, for example, $n$ words of frequency $z$ related to history topic will not be more important than other random $n$ words of the same frequency. At the same time, with `semantic fingerprints', topically connected words collectively increase or decrease expression of the corresponding semantic component or components: they are more important in determining the resulting fingerprint, than random noise words, even if they are frequent. This leads to better discrimination between documents of different topics. 

It is also important that `semantic fingerprinting' is significantly faster than `matrix translation', as we eliminate the necessity to look for the most similar neighbors of the predicted vector. This operation can be computationally expensive, especially on models with large vocabulary.

Note that one can apply the described method not only to proper cross-lingual translations, but also to problems like `projecting' texts in one style or genre into another. In fact the method is applicable to any situation, where there are two comparable sets of texts consisting of items, for which there theoretically exist pairwise links; one knows only a small part of these links, but would like to compare texts independent of the corpus they belong to. In these cases, semantic fingerprints method can be of use.

\section{Conclusion and Future Work}\label{sec:conclusion}
Thus, we described an approach to build language-independent semantic representations of documents in multi-lingual comparable corpora. It was tested on a rather small task of clustering Ukrainian and Russian academic texts into 3 topics. However, the results seem very promising to us and we plan to continue working on the proposed method. The models trained on our corpora, the linear transformation matrix, the evaluation dataset we used and Python code to work with this data are available online\footnote{\url{https://cloud.mail.ru/public/Eune/tN7ssqtWj}}.

The initial motivation behind this work was to develop a system for automatic plagiarism detection for two closely related languages. A crucial component of this system is a preprocessing part, which is able to cluster texts according to their topics. We believe that this component eventually will make it possible to compare `semantic fingerprints' of the documents in order to determine possible plagiarized texts and to perform their further analysis.

One obvious disadvantage of the proposed method is the necessity to know in advance the desired number of clusters (topics in the text collection). We plan to experiment with approaches to determining the optimal number of clusters automatically. It poses serious problems in multi-lingual settings, as the algorithms will be biased to language-based clustering, not taking into account topical division. Thus, ways should be invented to cope with this bias especially when the number of topics is much higher than 3 (used in this research).

Another direction of future work is to compare our approach and bag-of-words representations after proper machine translation. The results are not obvious: on the one hand, MT directly casts texts into another language and that should be a difficult baseline to beat. On the other hand, as explained in Section \ref{sec:disc}, dense document representations like semantic fingerprints can possibly be more flexible in grasping document contents than words-based representations.

Finally, we plan to test the proposed method with other language pairs, especially typologically distant languages. Experiments in \cite{mikolov2013translation} suggest that as long as the languages possess the meaningful notion of lexical co-occurrence, genetic or typologic distances between them should not matter. However, this is still to be tested and proved. 

\section{Acknowledgments}\label{sec:acknowledgements}
We are grateful to the voluntary \textit{Dissernet} community for providing us with the data, support, and inspiration to start this work. 

\section{Bibliographical References}
\label{main:ref}
\bibliographystyle{lrec2016}
\bibliography{par_cluster}

\end{document}